\documentclass[10pt,twocolumn,letterpaper]{article}

\usepackage{cvpr}              

\usepackage[dvipsnames]{xcolor}

\definecolor{cvprblue}{rgb}{0.21,0.49,0.74}
\usepackage[pagebackref,breaklinks,colorlinks,citecolor=cvprblue]{hyperref}
\usepackage{multirow}
\usepackage{bbm}
\usepackage{algorithm}
\usepackage{algorithmic}
\usepackage{cite}
\def\paperID{*****} 
\def\confName{CVPR}
\def\confYear{2024}

\title{Blurry-Consistency Segmentation Framework with Selective Stacking on Differential Interference Contrast 3D Breast Cancer Spheroid}

\author{Thanh-Huy Nguyen, Thi Kim Ngan Ngo, Mai Anh Vu, Ting-Yuan Tu\\
IMBSL Lab, Department of Biomedical Engineering\\
National Cheng Kung University, Tainan, Taiwan ROC\\
{\tt\small harvey.nguyen0811@gmail.com}
}

\begin{document}
\maketitle
\begin{abstract}
The ability of three-dimensional (3D) spheroid modeling to study the invasive behavior of breast cancer cells has drawn increased attention. The deep learning-based image processing framework is very effective at speeding up the cell morphological analysis process. Out-of-focus photos taken while capturing 3D cells under several z-slices, however, could negatively impact the deep learning model. In this work, we created a new algorithm to handle blurry images while preserving the stacked image quality. Furthermore, we proposed a unique training architecture that leverages consistency training to help reduce the bias of the model when dense-slice stacking is applied. Additionally, the model's stability is increased under the sparse-slice stacking effect by utilizing the self-training approach. The new blurring stacking technique and training flow are combined with the suggested architecture and self-training mechanism to provide an innovative yet easy-to-use framework. Our methods produced noteworthy experimental outcomes in terms of both quantitative and qualitative aspects.
\end{abstract}    
\section{Introduction}
\label{sec:intro}

Recent advancements in the utilization of 3D multicellular tumor spheroids have significantly progressed research in oncogenesis and tissue engineering, enriching our understanding of these domains \cite{weiswald2015spherical}. While various software tools exist for analyzing spheroid behaviors using microscopic data, they are limited to extracting specific cell morphologies at specific time points during time-lapse observation \cite{cisneros2016novel, ducker2020semi, grexa2021spheroidpicker, ngo2023deep}. Differential interference contrast (DIC) microscopy, with its fluorescence-free and non-destructive detection advantages, emerges as a widely preferred approach for observing live cells and tracking their dynamic progression temporally \cite{ziv2007differential}.

\begin{figure}[t]
	\centering
	\includegraphics[width=1\linewidth]{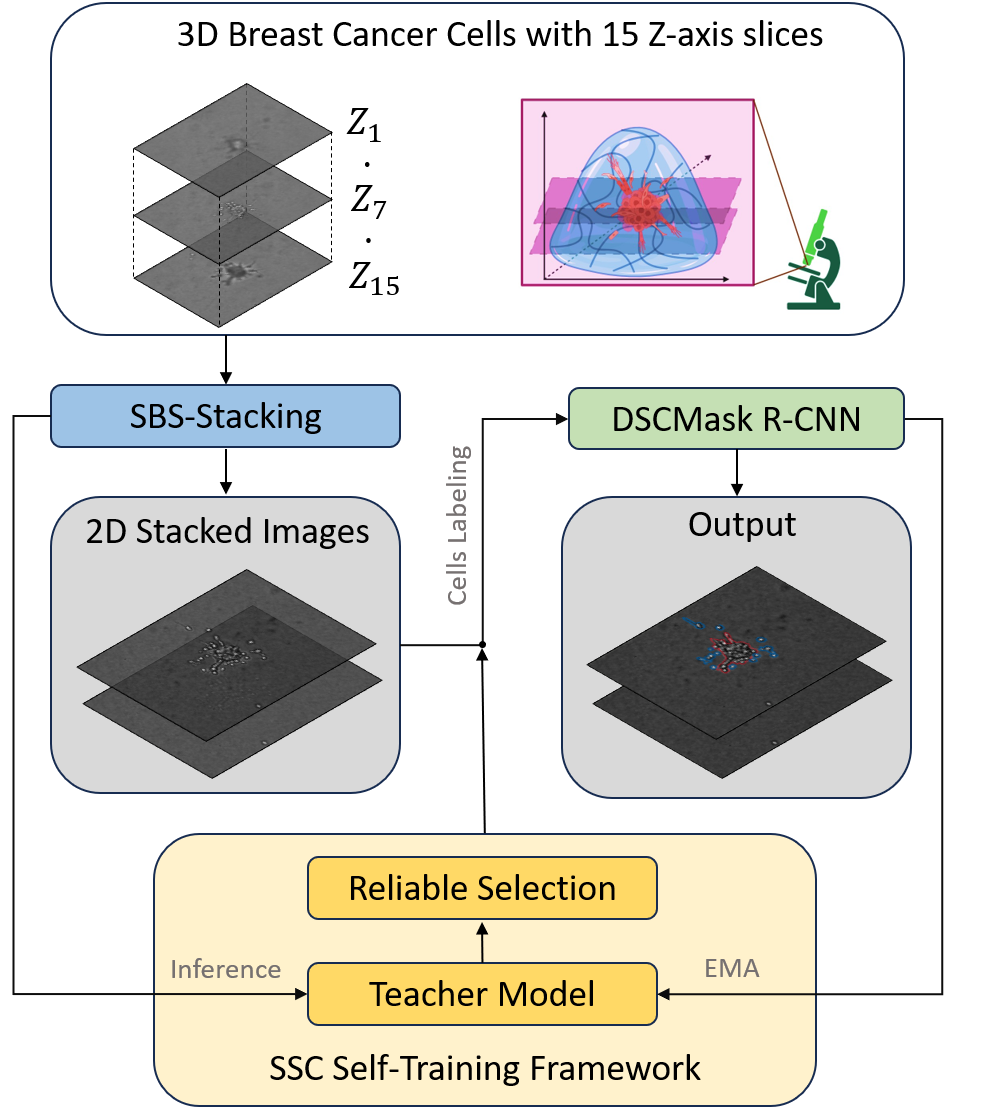}
	\caption{Our proposed pipeline for stacking and training 3D cell images. The 3D microscopy images are, firstly, stacked before running through the training stage of the proposed architecture DSCMask R-CNN. Finally, a self-training framework is employed to select reliable samples for the continuous training epochs.}
	\label{fig: 1}
\end{figure}
Despite its advantages, analyzing DIC spheroid images remains time-consuming and poses challenges in automating model extraction and analysis\cite{ngo2023deep}. The introduction of supervised machine learning frameworks, particularly deep convolutional neural networks (CNNs), has revolutionized medical image analysis \cite{nguyen2023context,truong2023delving,nguyen2023towards}. Various deep learning-based variants, such as encoder-decoder structures like U-net or two-stage instance segmentation like Mask R-CNN, leverage contextual information for image segmentation. Mask R-CNN model is an instance segmentation architecture leveraging the power of three heads, making them promising for analyzing spheroid behavior in DIC brightfield images.

Prior methods for 3D cell multiple slices have been extensively studied \cite{sigdel2015focusall}. 
Most of these methods use a variety of ways to transform the source image into different scales in an effort to improve resolution. However, traditional image quality enhancement from digital image processing approaches no longer shows good performance compared to deep learning approaches. The previous deep learning method \cite{tang2018pixel} learns focus measures to identify focused and defocused pixels in the source using an encoder-decoder-based approach. With the growth of consistency-based learning methods [], segmentation models have been assisted to perform excellently under noisy input, such as hard-to-learn samples. 

Having said that, most of those approaches are not well-utilized on 3D DIC microscope images due to the heavy blur effect of multiple slices. Focus Stacking itself showed promising performance with fast inference time and less sensitivity against the blur effect. In this study, our objective is to adapt the existing Focus Stacking framework, introducing a modified version termed Selective Blurry-Slice Stacking (SBS-Stacking) by filtering a high percentage of blurred areas to stack images before stacking. Besides, inspired by FixMatch \cite{sohn2020fixmatch}, we redesigned Mask R-CNN by adding a weak-to-strong augmentation mechanism to leverage the well-segmented sample from SBS-Stacking to teach fully-stacked blur images, called dense-stacked samples. Finally, to select useful information from stacked images with fewer slices, called sparse-stacked samples, we proposed a new self-training module. Different from dense-stacked information assisted by the consistency module, sparse-stacked information is utilized by a sparse-stacking consistency self-training framework that can select and prioritize reliable samples for re-training.

Our main contributions are summarized as follows:
\begin{itemize}
\item We proposed SBS-stacking methods to select high-quality slices for labeling and training procedures.
\item Inspired by semi-supervised-based pseudo-labeling methods, we redesigned Mask R-CNN that combined consistency regularization loss, called DSCMask R-CNN to leverage dense-stacked information from all slices.
\item We proposed a novel self-training scheme that utilizes sparse-stacked information that performs selective re-training via prioritizing reliable images.
\end{itemize}

\section{Related Works}
\label{sec:relatedworks}
\subsection{Biomedical Image Stacking}
Previous approaches for 3D cell multiple slices have been researched widely \cite{sigdel2015focusall}. 
These approaches fall into three categories: spatial domain (pixel values), transform domain (frequency components), and deep learning methods. Spatial domain image fusion selects pixels or regions from focused parts \cite{aslantas2014pixel}. Transform domain methods enhance resolution by transforming the source image into different scales, employing techniques like wavelet, curvelet, contourlet transforms, neighbor distance, Laplacian pyramid, or gradient pyramid \cite{kou2018multi}. Deep learning methods learn focus measures to identify focused and defocused pixels in source images \cite{tang2018pixel}, fusion operations to combine images without ground truth \cite{ram2017deepfuse}, and direct mappings between high and low-frequency images \cite{li2019multifocus}.

\subsection{Consistency Regularization}
Pseudo-labeling \cite{arazo2020pseudo} and consistency regularization \cite{bortsova2019semi,ngo2024dual} are two powerful strategies for exploiting unlabelled data. In the conventional pseudo-labeling method, the predicted pixels with high confidence are chosen as the pseudo-label for the unlabeled data using a high and defined threshold 
 \cite{lee2013pseudo}. However, only a small number of samples can surpass the selected threshold at the beginning of training with this technique. Consequently, Zhang et al. 
 \cite{zhang2021flexmatch} introduce the Curriculum Pseudo Labeling (CPL) technique, which modifies the flexible threshold of every category in real-time while it is being trained. Low-confidence pixels will nevertheless be eliminated in spite of the favorable results. Low-condence pixels are crucial for model training, as Wang et al. \cite{wang2022semi} showed. Consistency regularization seems to use data more, therefore this study will concentrate more on it. Consistency regularization concentrates more on how to obtain two identical predictions, such as data perturbation \cite{french2019semi}, model perturbation \cite{luo2021semi}, feature perturbation 
 \cite{ouali2020semi}, etc., even if it makes use of all available prediction data. However, when it comes to prediction optimization, a consistency loss function, such as L2 Loss, is used to treat all data consistently.

\subsection{Self-training on Semantic Segmentation}
The most basic pseudo-labeling and semi-supervised techniques are called self-training methods; they were initially introduced in \cite{yarowsky1995unsupervised}, extensively examined in \cite{triguero2015self}, and used with deep neural networks for the first time in \cite{lee2013pseudo}. These techniques involve feeding back the training set with the model's predictions, so retraining a base supervised model. It has been receiving more and more attention lately from a variety of domains, including domain adaptation \cite{zou2018unsupervised}, semi-supervised learning \cite{cascante2021curriculum}, and fully-supervised image recognition \cite{radosavovic2018data}. Specifically, it has been reconsidered in several semi-supervised tasks, such as semantic segmentation \cite{yuan2021simple}, object detection \cite{sohn2020simple}, and picture classification \cite{cascante2021curriculum}. J.Yuan et al. \cite{yuan2021simple} base their method on the premise that excessive data augmentations are detrimental to clean data distribution. Several methods for integrating data augmentation techniques into the self-training process have also been suggested. During the self-training phase, the ST++ \cite{yang2022st++} method uses data augmentation techniques on the unlabeled images.
\section{Methodology}
\label{sec:Methodology}
\subsection{Selective Blurry-Slice Stacking (SBS-Stacking)}
\label{sec:SBS-Stacking}
First, we derive a blur map for an image $\mathbf{B}$, characterizing spatially varying blur by modeling it as a composite of a sharp image ($\mathbf{S}$) convolved with a blur kernel ($\mathbf{b}$) and the addition of noise $\mathbf{n}$, as demonstrated in\cite{Golestaneh2017hifst}. 

\begin{equation}
\mathbf{B} = \mathbf{S} \odot \mathbf{b} + \mathbf{n}
\end{equation} 

The blur detection methodology involves the following key steps: 

\textbf{Gaussian Filtering for Noise Reduction}:  We employ a Gaussian filter, characterized by a small kernel function (g(x, y)), applied individually to each pixel (x, y) of the blurry image (B). The Gaussian filter is an objectives function defined by: 
\begin{equation}
g(x,y) = \frac{1}{{2\pi {\sigma ^2}}}{e^{ - \frac{{{x^2} + {y^2}}}{{2{\sigma ^2}}}}}
\end{equation} 

This process effectively mitigates high-frequency noise (introduced by the device lens). The resulting filtered image is denoted as Bg, retaining the image's inherent shape and structure while eliminating spatial redundancy. In the experiment setting, we chose a small $\sigma = 0.5$.

\textbf{Gradient Magnitude Calculation}: Let $B_g$ represent the input blurry image after it has been Gaussian filter. Convolution operations with horizontal ($h_x$) and vertical ($h_y$) gradient operators are applied to the Gaussian-filtered image ($B_g$) to compute the gradient magnitudes. The magnitude (G) is the square root of the sum of squared gradients. The  and $h_x$ and $h_y$ are defined as: 
\[{h_x} = \left[ {\begin{array}{*{20}{c}}
1&0\\
0&{ - 1}
\end{array}} \right],\,\,\,{h_y} = \left[ {\begin{array}{*{20}{c}}
0&1\\
{ - 1}&0
\end{array}} \right]\] 
Then, the gradient magnitude image G is calculated as:
\begin{equation}
\label{eq3}
G = \sqrt {{{\left( {{B_g} \odot {h_x}} \right)}^2} + {{\left( {{B_g} \odot {h_y}} \right)}^2}} ,
\end{equation} 

These steps collectively yield a comprehensive blur map reflecting the spatially varying blur in the image, providing valuable insights into the localized variations in sharpness.

Next, we apply HiFST\cite{Golestaneh2017hifst} decomposition on $G$ - the gradient-computed sample; blur can be interpreted differently depending on the scale. Images are processed with patch size M x M for each pixel centered at $(i,j)$.  Given that $H_{i,j}^M$ and ${L_{i,j}}$ are, respectively, a vector consisting of the absolute values of the high-frequency Discrete Cosine Transform (DCT) coefficients and its sorted vector. For each element $t$, ${L_{i,j;t}}$ is the $t$-th element of in vector $L_{i, j}$.  The matrix $L_{t}$ is defined as the ${N_1} \times {N_2}$ matrix defined as:  
\begin{equation}
{L_t} = \left\{ {{L_{i,j;t}},0 \le i < {N_1},0 \le j < {N_2}} \right\}
\end{equation} 

\begin{algorithm}[ht]
    \caption{Selective Blurry-Slice Stacking \\(SBS-Stacking)}\label{SBS-Stacking}
    \begin{algorithmic}
    \STATE \textbf{Input:} $B$: Blur images; $k$: top-k images that in-focus; $Z$: total number of slices.
    \STATE \textbf{Output:} $S:$ Sharp image.
        \FOR{$z = 1$ to $Z$}
            \STATE Calculate gradient magnitude image $G_{z}$ using Eq.\ref{eq3}
            \STATE $I_{z} = HiFST(G_{z})$ \COMMENT{Compute blur detection map.}
            \STATE $I_{BM}^z = \sum\limits_{i = 0}^{{N_1}} {\sum\limits_{j = 0}^{{N_2}} {{p_{i,j}}} } \,\,,{p_{i,j}} \in \left[ {0,255} \right] $ \{Compute in-focus pixel values of each $z$ image.\}
            \STATE $I_Z \leftarrow I_{BM}^z  $
        \ENDFOR
        \STATE Sort $I_Z$ by descending order.
        \FOR{$i = 0$ to $k-1$}
            \STATE ${I_{BM}^k}\leftarrow I_{BM}^z[i]$ \{Select top-k slices.\}
        \ENDFOR
        \STATE $S = FocusStack\left( {I_{BM}^k} \right)$
    \STATE \textbf{Return:} $S$.
    \end{algorithmic}
\end{algorithm}
 HiFST comprised of $\frac{{{M^2} + M}}{2}$ normalized layers $\widehat {{L_t}},\,\,1 \le t \le \frac{{{M^2} + M}}{2}$, where $\widehat {{L_t}}$ is shown as:
\begin{equation}
\widehat {{L_t}} = \frac{{{L_t} - \min \left( {{L_t}} \right)}}{{\max \left( {{L_t}} \right) - \min \left( {{L_t}} \right)}},\,\,1 \le t \le \frac{{{M^2} + M}}{2}
\end{equation} 
\\

Each layer can more effectively distinguish between the image's blurry and unblurred areas and measure the amount of blur locally by normalizing each layer between $\left[ {0,1} \right]$.

A blur detection map $D$ uses the first  $\sum _{r = 1}^m{M_r}$ layers is computed as:
\begin{equation}
D = T \circ \omega 
\end{equation}
whereas $\circ$ is Hadamard product and $M_{r} = 2^{2+r}$ if even and $M_{r} = 2^{2+r}-1$ if odd. Given that Matrices $T$ with ${0 \le i < {N_1},0 \le j < {N_2}}$, $T$ and $\omega$ elements are defined as:
\begin{equation}
{T_{i,j}} = \max \left( {\left\{ {\widehat {{L_{i,j;t}}}:t = 1, \ldots ,\sum\limits_{r = 1}^m {{M_r}} } \right\}} \right),
\end{equation}
\begin{equation}
{\omega _{i,j}} =  - \sum\limits_{\left( {i',j'} \right) \in R_{\left( {i,j} \right)}^k} {P\left( {{T_{i',j'}}} \right)\log \left[ {P\left( {{T_{i',j'}}} \right)} \right]} ,
\end{equation}
whereas ${R_{\left( {i,j} \right)}^k}$ is $k \times k$ patch centered at pixel $(i,j)$ and $P$ is probability function. After choosing $k=7$ for computing $7 \times 7$ neighborhood for the corresponding pixel in $T$, the entropy map $\omega$ is used as a weighting factor to give more weight to the microscopy image's salient regions. To reduce the impact of outliers and preserve boundaries, the final blur map is smoothed with edge-preserving filters.

After achieving a full saliency map of cell images, it can be seen that some of the z-slices have a high percentage of white area, whereas the blurry slices only contain most of the black area. Many previous similar approaches\cite{raudonis2021fast} consider stacking all the slices, which include many noise-contained slices. The output of the focus stacking algorithm is not well-represented for identifying among single cells. We propose the SBS-Stacking algorithm that aims to select high in-focus ratio slices to stack based on the achieved saliency map of each picture.

\begin{figure*}[t]
	\centering
	\includegraphics[width=1\linewidth]{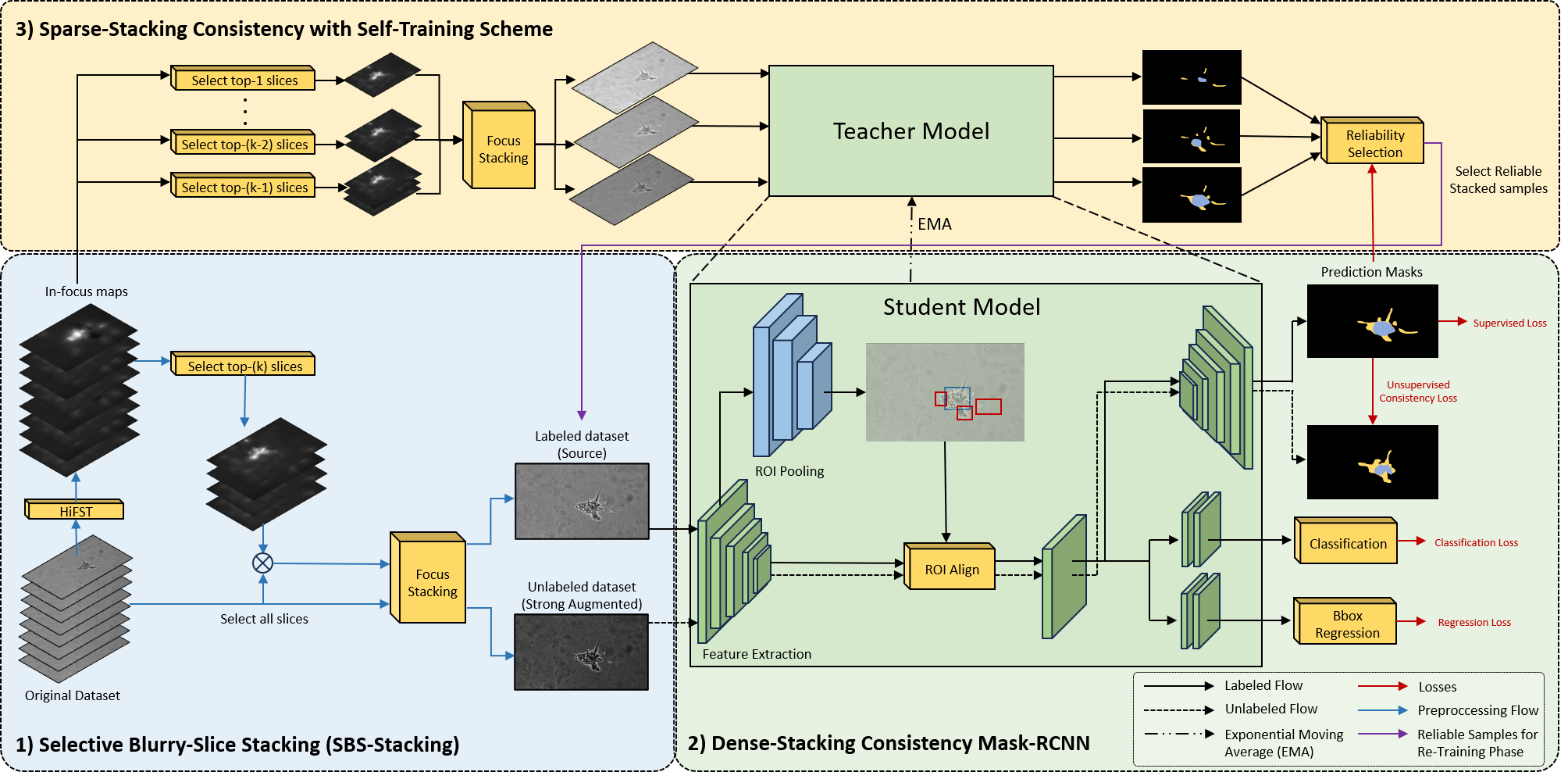}
	\caption{Proposed pipeline in detail with two main parts. 1) SBS-Stacking takes the original dataset to create partially stacked images and full focus-stacked images. 2) DSCMask R-CNN architecture that utilizes Mask R-CNN and mechanism of consistency training by treating SBS-Stacking images as a weak-augmented set and Focus-Stack images as a strong-augmented set. 3) Sparse-stacking consistency with a Self-training scheme will filter and select reliable images for the re-training stage by using a teacher model. 
 \vspace{5mm}}
	\label{fig: 2}
\end{figure*}

Given that, for every $z$-slice in $Z$, cumulative in-focus pixel values of an input image are computed by using estimated blur map $I_{BM}$ with the shape of $N_1 \times N_2$ as follows:
\begin{equation}
I_{BM}^z = \sum\limits_{i = 0}^{{N_1}} {\sum\limits_{j = 0}^{{N_2}} {{p_{i,j}}} } \,\,,{p_{i,j}} \in \left[ {0,255} \right]
\end{equation}
where $p_{i,j}$ is intensity value at pixel $(i,j)$. The k-selected slice with high in-focus pixel intensity is defined as:
\begin{equation}
I_{BM}^k = \left\{ {to{p_k}\left( {\left. {I_{BM}^z} \right|z \in Z} \right),0 < k \le Z} \right\}
\end{equation}

After selecting the k in-focus slice, the image output is focus-stacked as below:
\begin{equation}
S = FocusStack\left( {I_{BM}^k} \right)
\end{equation}
where FocusStack is an algorithm with the flow of using SIFT for feature detection, slices alignment with kNN feature matching, generating Gaussian blur, and Laplacian computing for combining multiple slices in the Z-axis.

The pseudo-code of the proposed stacking framework is shown in Algorithm \ref{SBS-Stacking}.

\subsection{Dense-Stacking Consistency Mask R-CNN (DSCMask R-CNN)}
\label{sec: DSCMask R-CNN}

In our work, we utilize the architecture of Mask-RCNN with a strong-weak consistency mechanism under multiple-slice stacking methods. For the input, the labeled set is the output of SBS-Stacking with $k$ slices, which plays a role as weak-augmented flow. Conversely, an unlabeled dataset, generated from all $z$ slices stacked, can be seen as a strong-augmented set.

For supervised training with a labeled set, we keep the original loss of Mask-RCNN, which includes classification loss, regression loss, and supervised mask loss. The Mask-RCNN loss is computed as:
\begin{equation}
{L_{MaskRCNN}} = {L_{cls}} + {L_{box}} + {L_{mask}}
\end{equation}
where classification loss is log loss, bounding box loss is regression smooth $L1$ loss, and mask loss is binary cross entropy loss. 

For consistency training with both datasets, we leverage the semi-supervised mechanism that uses a labeled set predicted mask as ground truth to teach unlabeled dataset pseudo masks. The consistency loss is cross-entropy loss, defined as:
\begin{equation}
{L_u} = \frac{1}{{{B_u}}}\sum \mathbbm{1}\left( {\max \left( {{p^\omega }} \right) \ge \tau } \right)H\left( {{p^\omega },{p^s}} \right),
\end{equation}

where $B_u$ denotes the batch size of unlabeled set, $\tau$ is a pixel-level threshold, and $p$ is pixel-level predicted probability. The overall objective function is a combination of supervised loss $L_{MaskRCNN}$ and unsupervised loss $L_u$ as:
\begin{equation}
L = {L_{MaskRCNN}} + \lambda{L_u}
\end{equation}
where $\lambda$, the relative weight of unsupervised loss, is a scalar hyperparameter.

\subsection{Sparse-Stacking Consistency with Self-Training Scheme}
\label{sec: SSC}

Based on our earlier research, which assumed that the SBS-Stacking method worked well enough, we discovered that tuning $k$-slice for training is not always stable and produces decent results. The reason for this is that the model is not consistent enough as a result of some samples lacking or sparsely distributed slices. We further employed the mechanism that post-selects certain samples with only a few slices yet is trustworthy enough for supervised training after stacking, called the Sparse-Stacking Consistency Self-training framework. Our goal is to develop a selective re-training technique that identifies the trustworthy stacked samples so that lower-slice samples can be safely exploited.

Previous works estimate the uncertainty or reliability of an image or pixel from various angles: for example, they train two differently initialized models to predict the same unlabeled sample and reweight the uncertainty-aware loss with their disagreements [17]; they also take the final softmax output as the confidence distribution and filter low-confidence pixels by pre-defined threshold [47, 63]. We want to test the dependability with a single training model in our proposed self-training framework, eliminating the need to manually select the confidence threshold. Additionally, we eliminate untrustworthy samples using image-level information as opposed to the often-used pixel-level information for a more steady reliability rating. During training, the model is also able to pick up more comprehensive contextual patterns because of the image-level selection.

In our work, mean IOU is employed to measure the stability and reliability among pseudo-masks from the student model and predicted masks from the teacher model. Given that, with any $n \in \left[ {0;k - 1} \right]$, considering stacked image $u_i^n \in {D^{{u^n}}}$, The meanIOU can serve as a measurement for stability and further reflect the reliability of the unlabeled image along with the pseudo mask:
\[s_i^n = \sum\limits_{j = 1}^{K - 1} {meanIOU\left( {M_{ij}^n,M_{iK}^n} \right)} \]
whereas $s_i^n$ is the stability value of an n-stacked image, showing the reliability of $u_i^n$. $K$ is the checkpoint during the training stage of DSCMask-RCNN. $M$ is the pseudo-mask predicted from the teacher and student model. After calculating each unlabeled image's stability score, we use these scores to order the entire n-slice stacked collection and compare it with the pre-defined threshold. In our experiments, the threshold $\tau$ for determining reliable sample shown as:
\[{r_i} = \left\{ {\left. {u_i^n} \right|s_i^n > \tau } \right\}\]
We set $\tau = 0.8$ for all of our experiments.
\section{Experiments and Results}
\label{experiments}

\subsection{Dataset}
Over the course of four days, the breast cancer cell line MDA-MB-231 was used to generate spheroids in a microwell dish. Following that, the spheroids were harvested and embedded in a collagen type I gel to allow for a 3D invasion assay. Continuous monitoring was accomplished using differential interference contrast (DIC) microscopy, which captured time-lapse images at 15-minute intervals over a 28-hour period using an objective lens with a magnification factor of 20X. A series of 15 z-stack images were acquired at each time point, spanning from the bottom to the top of the spheroid invasive region, with a consistent inter-plane distance of 20 micrometers. The collected images were then analyzed further to determine the invasive dynamics and behavior of the spheroid. In total, after stacking, three experienced annotators labeled 426 stacked images from 6390 slices with three classes: Invasive area, Spheroid core, and Single Cells.
\subsection{Implementation Details}
Our experimental settings are based on PyTorch using the NVIDIA RTX 2080Ti 11GB VRAM. For the preprocessing stage, we cleaned some bad samples by hand, and then SBS-Stacking was performed to stack $k$ images. We choose $k=8$, which is nearly half of the total 15 slices for every batch. For the training stage, our DSCMask R-CNN is built on top of the Mmdetection library. It was trained with 100 epochs, using an Adam optimizer with an initial learning rate of 0.001, weight decay of 1e-4, and a decay rate of 0.1 to minimize the total loss. Basic geometric augmentation is brought to reduce the model's bias, including rotation, flipping, and random cropping.  We set fixed $\lambda = 0.1$ to all experiments for consistency training with an unlabeled set.
\subsection{Qualitative Result: SBS-Stacking}

\begin{figure}[ht]
	\centering
	\includegraphics[width=1\linewidth]{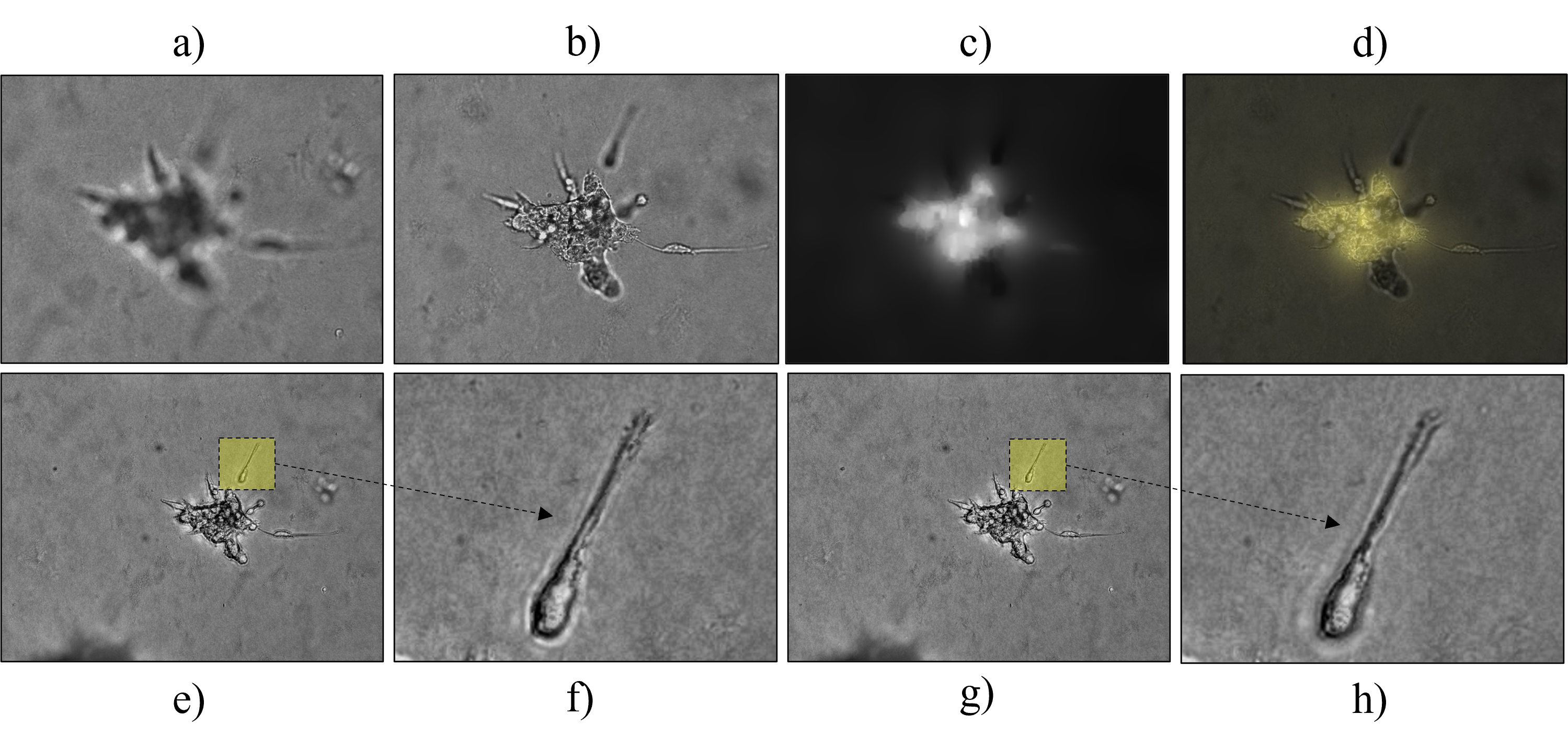}
	\caption{Visualization of a) Out-of-focus slice ($Z$-13); b) Partially in-focus slice ($Z$-6); c) and d) In-focus detection map of image b) and its overlay image; e) and f): Output images of SBS-Stacking; g) and h): Output images of normal Focus-Stack.  \vspace{3mm}}
	\label{fig: 3}
\end{figure}

Based on Figure \ref{fig: 3}, it can be seen that some of the slices in the cell image are out-of-focus and full of blurred parts, as image \ref{fig: 3}a. Meanwhile, image \ref{fig: 3}b contains more sharp areas but still has many blurry areas. Through the HiFST algorithm, image  \ref{fig: 3}c shows its in-focus area with a grey area, which overlaps nearly 80\% of the cells. In the stacked images, our proposed SBS-Stacking shows its potential with the sharp contours and less effect of blurry slices, as in image \ref{fig: 3}e and its zoom-in image \ref{fig: 3}f. The traditional Focus-Stack without cleaning blurry images shows the noisy pixels with no clearly sharp boundary of cells, as in the image 
 \ref{fig: 3}g and image \ref{fig: 3}h.
\subsection{Quantitative Result: DSCMask R-CNN}
\begin{table}[ht]
\centering
\label{table1}
\renewcommand{\arraystretch}{1.5} 
\resizebox{1\columnwidth}{!}{%
\begin{tabular}{|c|c|cc|cc|}
\hline
\multirow{2}{*}{}             & \multirow{2}{*}{Backbone} & \multicolumn{2}{c|}{Bounding box evaluation}             & \multicolumn{2}{c|}{Mask evaluation}                    \\ \cline{3-6} 
                              &                           & \multicolumn{1}{c|}{mAP@{[}0.5:0.95{]}} & mAP@0.5        & \multicolumn{1}{c|}{mAP@{[}0.5, 0.95{]}} & mAP@0.5        \\ \hline

\multirow{4}{*}{Ngo TKN \cite{ngo2023deep}}   & R50                 & \multicolumn{1}{c|}{-}              & -          & \multicolumn{1}{c|}{0.559}             & 0.832          \\ \cline{2-6} 
                              & R101                & \multicolumn{1}{c|}{-}              & -         & \multicolumn{1}{c|}{0.571}             & 0.846          \\ \cline{2-6} 
                              & X101-32x4d                & \multicolumn{1}{c|}{-}              & -          & \multicolumn{1}{c|}{0.579}             & 0.852          \\ \cline{2-6} 
                              & X101-64x4d                & \multicolumn{1}{c|}{-}              & -         & \multicolumn{1}{c|}{0.586}             & 0.868          \\ \hline

\multirow{4}{*}{Mask R-CNN}   & R50 - FPN                 & \multicolumn{1}{c|}{0.690}              & 0.930          & \multicolumn{1}{c|}{0.620}             & 0.918          \\ \cline{2-6} 
                              & R101 - FPN                & \multicolumn{1}{c|}{0.702}              & 0.946          & \multicolumn{1}{c|}{0.631}             & 0.932          \\ \cline{2-6} 
                              & X101-32x4d                & \multicolumn{1}{c|}{0.716}              & 0.958          & \multicolumn{1}{c|}{0.642}             & 0.947          \\ \cline{2-6} 
                              & X101-64x4d                & \multicolumn{1}{c|}{0.724}              & 0.962          & \multicolumn{1}{c|}{0.650}             & 0.959          \\ \hline
\multirow{4}{*}{DSCMask R-CNN} & R50 - FPN                 & \multicolumn{1}{c|}{0.692}              & 0.938          & \multicolumn{1}{c|}{0.628}             & 0.931          \\ \cline{2-6} 
                              & R101 - FPN                & \multicolumn{1}{c|}{0.704}              & 0.944          & \multicolumn{1}{c|}{0.636}             & 0.940          \\ \cline{2-6} 
                              & X101-32x4d                & \multicolumn{1}{c|}{0.722}              & 0.960          & \multicolumn{1}{c|}{\textbf{0.654}}    & 0.964          \\ \cline{2-6} 
                              & X101-64x4d                & \multicolumn{1}{c|}{\textbf{0.730}}     & \textbf{0.966} & \multicolumn{1}{c|}{0.652}             & \textbf{0.967} \\ \hline
\end{tabular}
}
\caption{Quantitative results of Mask R-CNN and DSCMask R-CNN with various backbones compared to previous related work on stacked 3D cells dataset.}
\end{table}

In this experiment, we compared our methods with previous related work. Similar to our current work, Ngo TKN et.al. \cite{ngo2023deep} work focuses on 2D modality breast cancer cells. The work is based on the Unet++ semantic segmentation network that does not perform bounding box prediction as our two-stage approach. For a fair comparison, we implemented their methods under similar backbones. Their works proposed multiple models for performing segmentation on invasive areas, single cells, and core areas. However, we only used the invasive areas model for comparison with our methods. The hyperparameter config was set similarly to our proposed architecture.

According to Table \hyperref[table1]{1}, the DSCMask R-CNN shows the overall improvement on all feature extractors. In our work, we use mAP as the main evaluation metric for both instance segmentation and detection. It can be seen that Mask R-CNN shows the highest performance on the setting of X101-64x4d with 72.4\% and 65.0\% with $mAP@[0.5,0.95]$. Our proposed DSCMask R-CNN utilizes the mechanism of consistency training achieved on both tasks, with 73.0\% and 65.2\% on $mAP@[0.5,0.95]$ respectively. For $mAP@0.5$, our best result achieved on both tasks is up to 96.6\% and 96.7\%, respectively. As a consequence, the stacking method for preprocessing might not always be ideal for all samples. Consistency training can reduce the significant amount of unclear boundary problems by leveraging weak-to-strong self-training. Compared to previous work, our method shows promising results against different approaches that use semantic segmentation with the U-net family. With similar backbones, DSCMask R-CNN architecture outperforms nearly 10\% with $mAP@0.5$
metric over previous work\cite{ngo2023deep}.
\subsection{Ablation Studies on Proposed Framework}
We compared multiple settings of our framework as ablation studies as shown in Table \hyperref[tab: ablation studies]{2}. We used mAP@[0.5,0.95] for our evaluation metric for mask prediction. It can be seen that the baseline model that uses the SBS-Stacking method and the Mask R-CNN model achieved 62.32\%, lower than 1,86\% compared to the setting that applied SSC Self-training. We changed Mask R-CNN into DSCMask R-CNN without SSC Self-training, which reached 63.92\%, higher than using Mask R-CNN 1.6\%. The combination of all methods reaches the peak at 66.52\%, which is our current state-of-the-art on the differential interference contrast 3D Breast Cancer Spheroid dataset.
\begin{table*}[ht]
\caption{Ablation studies of our three proposed modules using ResNet-50 RPN as backbone on 3D breast cancer spheroid dataset}
\label{tab: ablation studies}
\begin{center}
\begin{tabular}{l@{\hskip 0.2in}|c@{\hskip 0.2in}c@{\hskip 0.1in}c@{\hskip 0.1in}c@{\hskip 0.1in}|c@{\hskip 0.1in}c}

\toprule
 & SBS-Stacking &  Mask R-CNN  & DSCMask R-CNN & SSC Self-training & mAP@[0.5, 0.95] \\
\midrule
\midrule
Baseline & \checkmark& \checkmark  &  &  &  62.32\\
\midrule
\multirow{4}{*}{Proposed Methods} 
     & \checkmark& \checkmark  &  &  \checkmark &  64.18\\ \cmidrule(){2-7} 
     & \checkmark&   & \checkmark  &  &  63.92 \\ \cmidrule(){2-7}
     & \checkmark&   & \checkmark & \checkmark &  \textbf{66.52}& \\
\bottomrule
\end{tabular}
\end{center}
\end{table*}

\section{Conclusion}
Overall, our work aims to propose a new method to handle the 3D breast cancer cells DIC microscopy dataset by stacking them with SBS-Stacking. The algorithm eliminates the unclear boundary caused by blurry slices from the original batch of images. Besides, our combination Mask R-CNN with weak-to-strong consistency training, named DSCMask R-CNN, also helps the model avoid biases and boosts the performance of architecture under various backbones. Lastly, our proposed self-training framework has boosted the performance by selectively utilizing the fewer-slice image for the re-training stage. Our dense-stacked and sparse-stacked handling was approached in different ways using self-training and consistency regularization, respectively. In future work, we aim to expand from instance segmentation to tracking tasks. Biological analysis will be conducted from our robust framework.

\section{Compliance with ethical standards}
\label{sec:ethics}
The research data was conducted by the IMBSL Lab, Department of Biomedical Engineering, National Cheng Kung University. This study did not require any ethical approval.

{
    \small
    \bibliographystyle{ieeenat_fullname}
    \bibliography{main}
}


\end{document}